\documentclass[a4paper, 10pt, conference]{ieeeconf}      

\IEEEoverridecommandlockouts                              
\usepackage{graphicx} 
\graphicspath{ {./images/} }
\title{\LARGE \bf
Geographical Map Registration and Fusion of Lidar-Aerial Orthoimagery in GIS
}

\author{Siqi Yi, Stewart Worrall and Eduardo Nebot*
\thanks{*Authors are from Australian Center for Field Robotics, the University of Sydney, Australia}
}

\begin{document}

\maketitle
\thispagestyle{empty}
\pagestyle{empty}

\begin{abstract}
Centimeter level globally accurate and consistent maps for autonomous vehicles navigation has long been achieved by on board real-time kinematic(RTK)-GPS in open areas. However when dealing with urban environments, GPS will experience multipath and blockage in urban canyon, under bridges, inside tunnels and in underground environments. 
In this paper we present strategies to efficiently register local maps in geographical coordinate systems through the tactical integration of GPS and information extracted from precisely geo-referenced high resolution aerial orthogonal imagery.
Dense lidar point clouds obtained from moving vehicle are projected down to horizontal plane, accurately registered and overlaid on aerial orthoimagery. Sparse, robust and long-term pole-like landmarks are used as anchor points to link lidar and aerial image sensing, and constrain the spatial uncertainties of remaining lidar points that cannot be directly measured and identified. We achieved 15-75cm absolute average global accuracy using precisely geo-referenced aerial imagery as ground truth. This is valuable in enabling the fusion of ground vehicle on-board sensor features with features extracted from aerial images such as traffic and lane markings. It is also useful for cooperative sensing to have an unbiased and accurate global reference. Experimental results are presented demonstrating the accuracy and consistency of the maps when operating in large areas.
\end{abstract}

\section{INTRODUCTION}
Despite rapid progresses being made in the field of autonomous vehicles localization and mapping, reliable maps are generally made by either utilizing high precision GPS, or in local coordinate systems utilizing on-board relative sensors. Relative sensors accumulate error and drift proportionally to the distance travelled. Loop closures along the vehicle route can constrain the growing uncertainty, but the map still cannot be registered or consistent within a global coordinate system due to lack of global references. Furthermore, a single false positive loop closure is enough to cause catastrophic faults. These two fundamental issues make loop closures alone not a solution for robust mapping when working in large areas. 

Position information in a global coordinate frame can be obtained using GPS, and in certain circumstances this can be in the centimeter range using RTK corrections.
When operating in an urban environment, there are well known fundamental limitations to GPS localisation due to multipath effects, and in GPS-denied areas such as urban canyons, tunnels, underground or indoors. 
Furthermore, in relatively open sky areas, GPS accuracy also varies depending on satellite configuration and atmospheric conditions at the time of obtaining data. 
The result of this is an unobservable amount of error and bias to the estimated position.
Therefore it becomes essential to integrate all forms of GPS and other absolute information to on-board relative sensors to be able to localize reliably and with predictable level of accuracy in global frames, preferably frames consistent with geographical coordinate systems.

\begin{figure}
    \centering
    \includegraphics[width=\columnwidth]{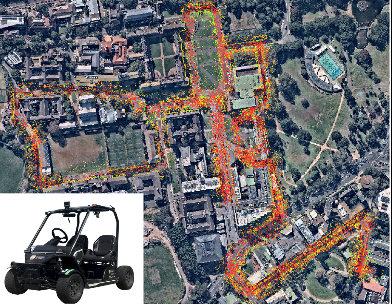}
    \caption{The lidar point clouds superimposed and projected on horizontal plane as a layer in GIS is precisely registered and overlaid onto high resolution orthogonal aerial imagery. The dataset was taken by an electric vehicle starting from bottom right corner, driving through the University of Sydney main campus in multiple loops, and back to the starting location. The electric vehicle is equipped with Velodyne VLP 16 beam lidar and multiple sensors}
    \label{fig:full_overlay}
\end{figure}

This paper proposes a strategy to georeference a locally consistent map of lidar based features by first incorporating correspondences with low quality GPS information (when available), and finally to integrate high quality feature observations derived from high resolution aerial imagery. This approach considers the limitation of GPS but still makes use of this information to assist with the global registration. 
It is well known that standard GPS accuracy in an urban environment is a minimum of 3 to 5 meters, with statistical outliers and bias commonly observed.
Despite this, the resulting position and heading are reliable and consistent at large scale and over long distances.
We take advantage of this fact in the registration strategies described in section \ref{rigid} and \ref{loose}.   
To fine-tune the map, we also present a method in section \ref{anchor} that utilizes manually labeled landmarks identified from high resolution aerial orthoimagery as reference points to anchor other map features whose geographical positions cannot be directly measured.   

Long term stable, robust, sparse landmarks such as poles and building corners are ubiquitous in urban environments in any cities around the world. In section \ref{landmarkslamrelatedwork} a number of recent works in landmark localization are reviewed. However there are areas that landmarks are too sparse to provide continuous update for localization. In these areas, dead reckoning and other external sensing capabilities are needed to fill in the gap. In our previous work\cite{SiqiMetrics}, metrics for evaluating update intermittence is presented and directly related to localization robustness. In the scenario of mapping and sensor fusion, high level robust landmarks such as poles can serve as fusion anchors across sensor modalities. These type of landmarks have been the object of significant research in previous years. There are very robust detection algorithms that can work for a variety of landmarks and different sensor modalities, including laser, vision etc. These algorithms have utilised lidar pointclouds, semantic segmentation, monocular and stereo vision. In this paper we explore the use of pole-like landmarks as anchors to minimize inconsistencies and biases between lidar features and aerial images. Vehicle trajectories, and dense lidar pointclouds can then be accurately represented in the global reference frame. 

There are a number of benefits for using geographically accurate landmarks. 
\begin{itemize}
\item Simplified management of large scale maps utilizing existing geographic coordinate systems and geographic information system(GIS) databases. This eliminates the need for complex sub-maps and hierarchical map structures.
\item Accumulated error resulting in large scale drifts can be corrected by incorporating observations from a global coordinate frame.
\item Global observations such as from GPS can be integrated within an online localization and sensor fusion framework. When available, GPS can be used as a coarse pose prior for initialization and relocalization against a georeferenced feature map.
\item Registration and fusion of any on-board sensor information to GIS/aerial imagery, as long as the sensors extrinsic parameters are known. For example, dense visual features, lidar pointclouds and other sensor information can be georeferenced.
\end{itemize}

\section{CONTRIBUTIONS}

The work proposed in this paper aims to build from existing work in graph slam by first building correspondences between features and using vehicle ego sensors to determine motion. The first contribution is the addition of the weak correspondences between the sparse GPS based locations (from the filtered output which rejects outliers) and the graph slam which is then optimised. The result of this is a globally referenced map that is reasonably consistent over a large and has error proportional to the quality of the GPS - including bias, etc.  The second contribution looks to georeference this information relative to aerial based imagery. Features are selected (currently manually but future automatically) from the aerial images that have corresponding lidar features in the feature based graph described already. These correspondences are incorporated into the graph as tight constraints. The results show that only a small number of these are required to improve the overall quality of the position information for the entire map

This paper is organized as follow. Section \ref{relatedwork} presents an overview of existing work in geographical and landmark based map creation and localization. Detailed strategies and steps for registration methodology are described in section \ref{method}. Section \ref{result} presents quantitative evaluation and result of fusion accuracy. Section \ref{conclusion} concludes this paper with potential future directions.

\section{RELATED WORK} \label{relatedwork}
\subsection{Map making methodologies for robot navigation}
There are different categories of methodologies to create and adjust maps for the use of autonomous vehicle localization. 

First, the methods based on local frame of reference using relative sensors. These approaches are subject to integrated error and have only local consistency. They include Slam variants, visual odometry  and can be build with dense or feature based methods. Although made locally, some of slam works such as \cite{newman2006outdoor} manually scale and transform robot trajectories and overlay onto a satellite image to visually validate how well the trajectories follow road and building structures in public satellite images. 



Second, georeferenced maps made from combining high precision GPS with dense pointclouds 
or features.
\cite{wan2018robust} utilizes RTK-GPS which could work well in open areas but will experience difficulties in urban canyons scenarios, tunnels, and indoor car parks. 
There are also works using filters, inertial measurement unit(IMU), and other sensors to reckon vehicle poses during outage of GPS signal. \cite{qu2015vehicle} localized in urban Paris using traffic signs detected from geo-referenced images, but the image geo-registration process is done regardless of GPS outage and multipath by a navigation system fusing IMU, wheel odometer and differential GPS readings. 

Third, without the high precision GPS, other sources of global information have to be used to fit locally consistent maps to global reference. This includes overlaying to satellite, matching to city models, air-borne lidar pointclouds, and street maps. To correct for the scale and pose drift, \cite{lothe2009towards}\cite{lothe2010real} registered distorted visual slam map and robot trajectory by non-rigid ICP matching with coarse geo-referenced 3D model of the environment provided by GIS database. A second step then optimizes additional geometric constraints relative to the 3D model. However public 3D city models are hardly available and the precision is usually coarse. \cite{ventura2014global} globally localized within geo-registered lidar pointclouds using mobile phone cameras. UTM coordinates of airborne lidar point clouds are obtained from alignment to 2D polygonal building roof outlines. 
In \cite{guivant2007global}, particle filter localization exploits publicly available geo-referenced urban street networks, more specifically the route network description file (RNDF) as prior information. Local RNDF road segments are converted into grid based maps to constrain the likely particle assignments.
\cite{tournaire2006towards} uses zebra line road markings to finely geo-register images obtained from ground vehicles to aerial images. \cite{sun2016extraction} also extracted and reconstructed zebra crossings from aerial images. 
\cite{bodensteiner2015single} geo-registered video frames captured by quadcopters shared in community video websites. This is done by ICP matching visual slam feature pointclouds extracted from the videos to geo-refenced lidar pointclouds.



\subsection{Sparse high level landmarks as features for localization and mapping} \label{landmarkslamrelatedwork}
Sparse high level landmarks with distinct geometrical shapes have been widely used in the literature of localization and mapping. Pole like features are detected from on-board lidar\cite{kuhneraccurate}\cite{weng2018pole}\cite{schlichting2014localization}, stereo camera\cite{spangenberg2016pole}\cite{nedevschi2018method} and pole detection neural networks from monocular camera\cite{kampker2018concept}. \cite{sefati2017improving} showed its lidar pole detection is more precise than stereo-camera algorithm. \cite{shen2018conditional} presents a graph-based localization algorithm using only tree and pillar landmarks. A novel ICP algorithm is proposed for more efficient and accurate landmark data association. Building corners\cite{im2016vertical} and facades\cite{kuhneraccurate}\cite{bansal2011geo}\cite{schlichting2014localization}, are also exploited as localization landmarks or references. Lidar and camera can also detect and localize building windows\cite{wang2011window} and doors\cite{nguatem2014localization}. \cite{bansal2011geo} established correspondence of building facades between oblique aerial/satellite images and ground robots acquired images. 


\section{MAP GEOGRAPHICAL REGISTRATION} \label{method}

A topological map was generated using a graph-based slam algorithm incorporating correspondences between lidar based features matched using ICP, and vehicle odometry from an IMU and wheel encoders. Our operating domain is the University of Sydney, which is an urban type environment including tall buildings, open spaces, areas with narrow roads, an underground carpark and other GPS denied areas. 
The relative pose information was incorporated into a graph structure which was then optimised using the g2o library\cite{kummerle2011g}.
The proposed georegistration methods were applied to this graph structure, though it is possible to use these methods for other algorithms which provide a map that is only consistent in a local frame of reference, such as slam variants or visual odometry.


\subsection{Rigid alignment of local vehicle path to GPS vehicle path} \label{rigid}
In addition to the relative sensor information, standard quality GPS positions was also collected at 1Hz.
The locally consistent graph optimised vehicle path, and the path measured using the GPS appear to have a similar shape despite their own inaccuracies. They also have known one-to-one correspondences (though with error and bias) according to the sensor time stamps. We can directly find the optimal SE2 transformation by minimizing distances between relative vehicle poses and their corresponding GPS observations. The resulting transform is used to translate and rotate both the map of the features (landmarks) and vehicle path from vehicle relative coordinate system to a globally referenced coordinate system. 

The rigid alignment of the locally consistent map to the GPS position treats the relative map as a rigid body.
This is beneficial for a map covering a small area where the drift from accumulated error is not significant, or where the GPS is of poor quality. If GPS position information containing statistically significant outliers was incorporated directly into the graph optimisation, the result would be a negative effect on the overall map.
The approach of finding the rigid transform between the locally consistent graph and the GPS trajectory would reduce the influence of a small number of outliers.

\begin{figure}
    \centering
    \includegraphics[width=\columnwidth]{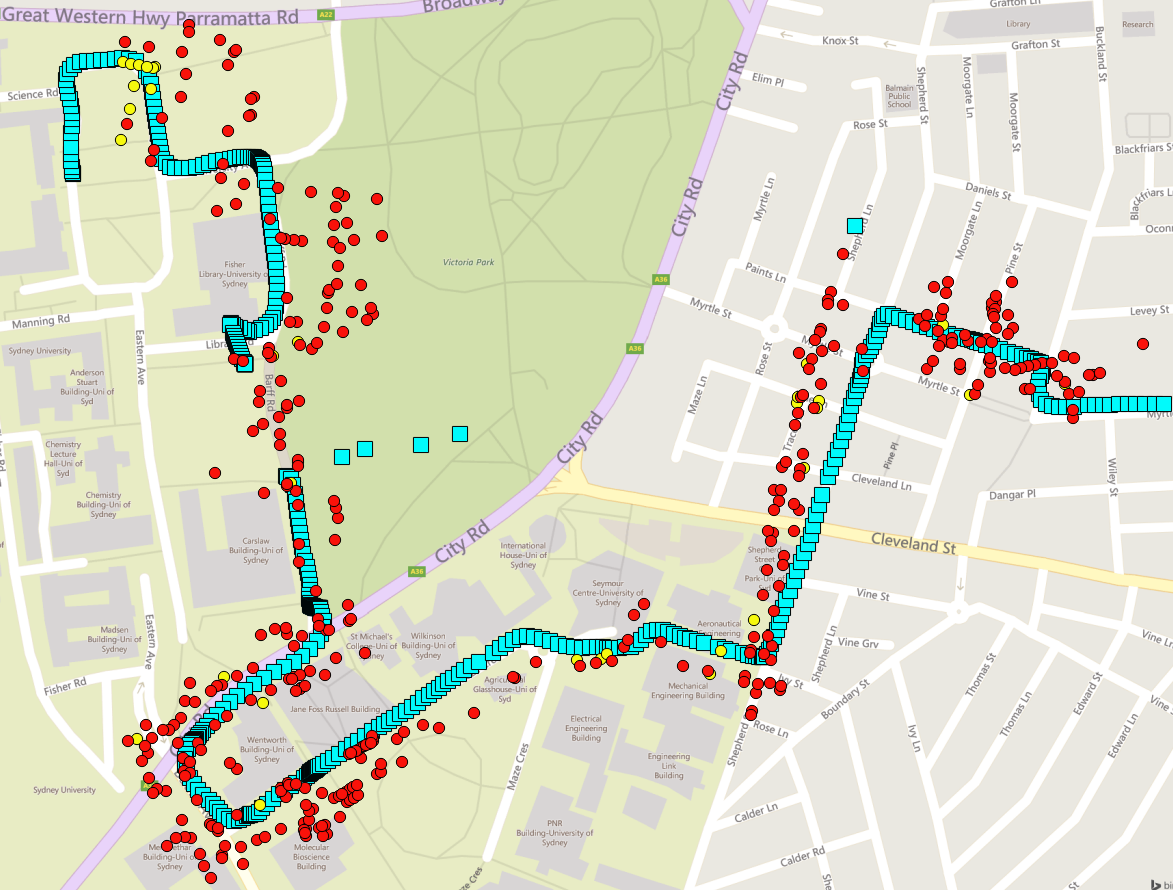}
    \caption{Result of rigidly aligned landmark map with GPS vehicle path overlaid on street map in GIS from method \ref{rigid}. Red and yellow dots are poles and building corners respectively. Blue squares are raw GPS readings along the vehicle driving path. Noise and discontinuity at the middle section are due to vehicle entering an underground carpark, where GPS readings drifted thousands of meters away. The registration is roughly correct regardless of GPS errors. Landmarks at both ends of driving path drifted due to accumulated uncertainty.}
    \label{fig:quadtoqueen_1}
\end{figure}

Figure \ref{fig:quadtoqueen_1} shows the resulting relative map of lidar landmarks overlaid on aerial map in GIS visualization software using this registration method. The middle section of path is aligned well while the beginning and end depart from their true locations due to accumulated relative sensor uncertainties.

\subsection{Loose GPS constraints} \label{loose}
\begin{figure}
    \centering
    \includegraphics[width=\columnwidth]{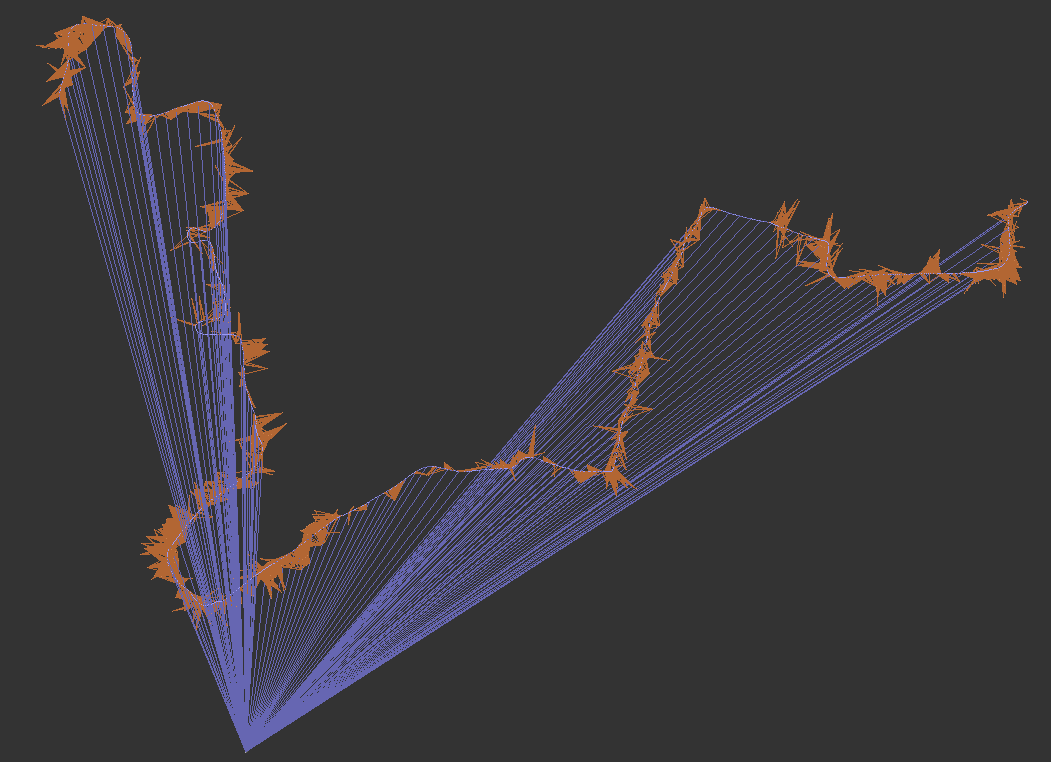}
    \caption{Graph structure of method \ref{loose}. Filtered GPS readings are added to the graph every 10 meters and represented in blue edges. All GPS edges have a global reference to map origin, which has a fixed offset and same heading to UTM and is randomly and conveniently chosen for visualization and numerical efficiency(smaller float numbers instead of UTM coordinates). Orange edges and remaining blue edges are lidar landmark observations and dead-reckoning edges obtained from IMU and wheel encoder, respectively. }
    \label{fig:graph}
\end{figure}
\begin{figure}
    \centering
    \includegraphics[width=\columnwidth]{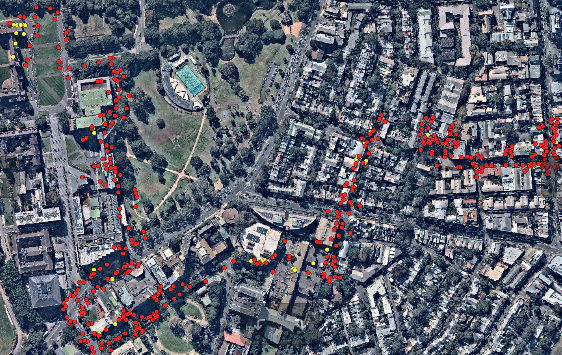}
    \caption{Resulting landmark map from method \ref{loose} overlaid on aerial map in GIS. Drifts are corrected and landmark positions follow road networks. The improvement is more noticeable at the beginning and end of the run when compared with previous approach.}
    \label{fig:quadtoqueen}
\end{figure}
In a graph-based slam framework, landmarks and vehicle poses are modeled as vertices connected by edges that encode relative sensor observations between poses. We added additional constraints to a limited number of vehicle pose vertices by incorporating edges representing GPS correspondences to a fixed map origin, see Figure \ref{fig:graph}.
Due to the high uncertainty and poorly defined error profile of the GPS information, these links are considered loose relationships with the corresponding uncertainty set very high. This high uncertainty for the GPS means that the existing relative graph relationships have a larger effect on a local scale in the resulting map, but the accumulated error of the relative information means that the GPS information will affect, and correct, the map on a global scale.

The map origin has a fixed spatial offset from origin of UTM zone 56S and no directional offset. It is chosen arbitrarily on campus for the convenience of visualization and numerical efficiency (smaller floating point numbers instead of large UTM coordinates). A global optimization process shifts all vertices in the graph to their optimal global poses with respect to map origin as this is the only fixed vertex. The map is then transformed from map coordinates to UTM frame by simply adding the Easting and Northing offset. The number of GPS constrained vehicle pose vertices and GPS edges required can vary according to GPS and local map quality. For the map in Figure \ref{fig:graph}, one GPS edge is added for every 10m of travel and GPS edges have a fixed high standard deviation of 5m in both Easting and Northing. The resulting landmark map is shown in Figure \ref{fig:quadtoqueen}. The results are now much more global acccurate as can be seen from beginning and end of the trajectory when compared to Figure \ref{fig:quadtoqueen_1} 

\begin{figure}
    \centering
    \includegraphics[width=\columnwidth]{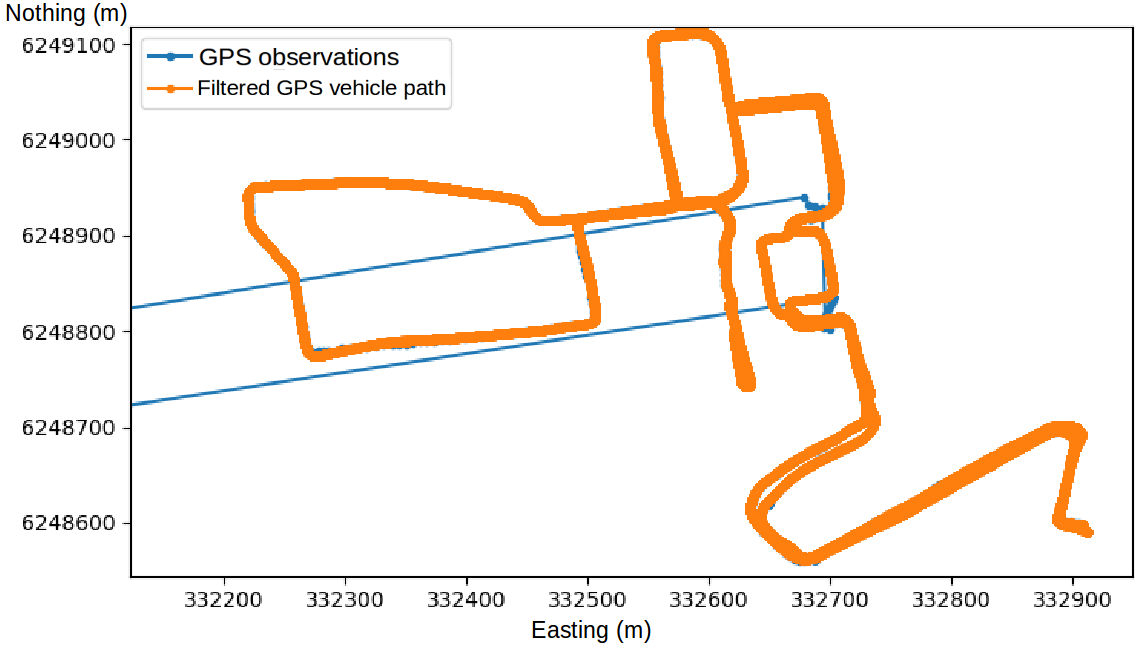}
    \caption{Raw GPS readings and UKF-filtered GPS vehicle path in blue and orange. Vertical and Horizontal axis are Northing and Easting in UTM zone 56S.}
    \label{fig:gps_path}
\end{figure}
\begin{figure}
    \centering
    \includegraphics[width=\columnwidth]{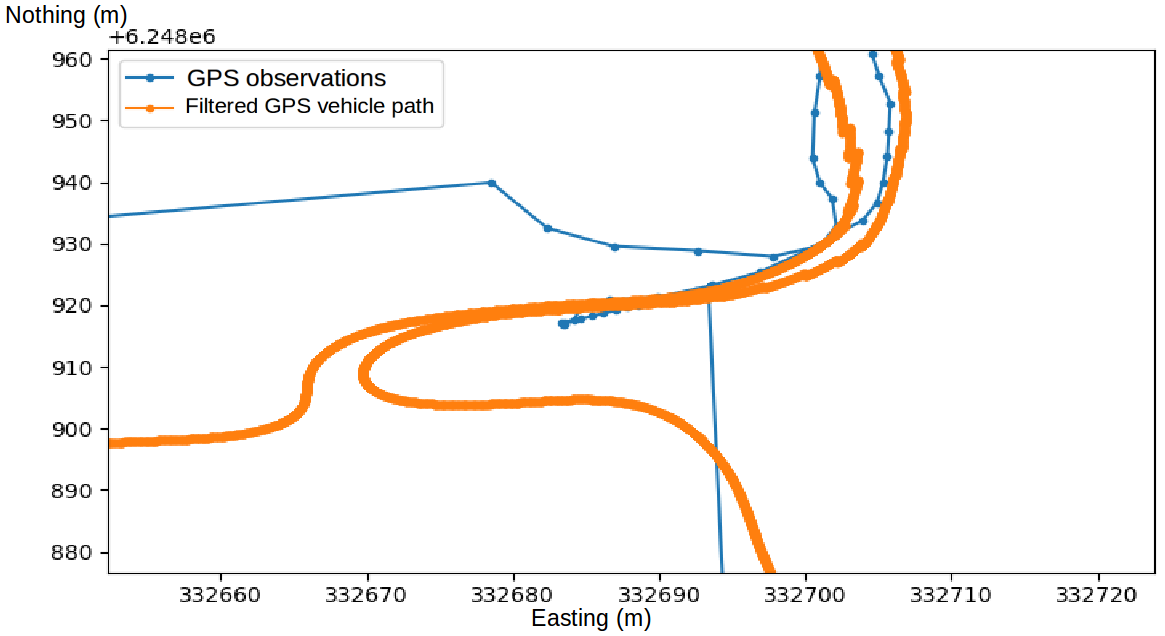}
    \caption{Close-up look at entrance of underground carpark. As GPS signal qA blocked underground, GPS readings became extremely unreliable(blue dots go far away and out of scope of this picture). Vehicle rejected these GPS outliers when driving in underground carpark and relied on only dead reckoning. After coming out of underground, GPS signals were in the proximity of vehicle states again, so vehicle resumed incoporating GPS updates.}
    \label{fig:gps_carpark}
\end{figure}
\begin{figure}
    \centering
    \includegraphics[width=\columnwidth]{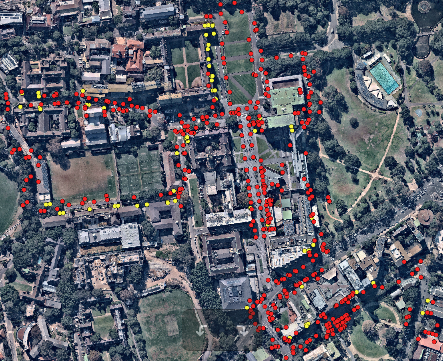}
    \caption{Resulting landmark map from method \ref{loose} overlaid on aerial map in GIS. Drifts are constrained by both GPS edges and multiple loop closures.}
    \label{fig:map}
\end{figure}

When vehicles do not revisit the same areas multiple times, there is no opportunity to constrain the growing uncertainty using loop closure making the resulting trajectory more susceptible to drift from accumulated error. 
This deviation is further amplified by heading errors, as small heading errors are magnified as spatial drifts over longer distances. 
Although it is difficult to model the error profile of a GPS at a local scale, the uncertainties and biases do not accumulate over distance. 
This means that the additional constraints of the GPS can correct for drift over a large map.
As the GPS observations will warp the map towards the globally consistent position, it is necessary to filter the GPS outliers to prevent potentially severe distortions of the locally consistent map. 
Figure \ref{fig:gps_path} shows raw GPS readings alongside a UKF-filtered vehicle path of drive around the campus. 
The GPS coordinates are represented in EPSG:32756 WGS 84/UTM zone 56S. 
The UKF makes predictions using the IMU and wheel encoder, and uses observations from the GPS to make updates. 
The filter also rejects GPS outliers that fall outside 95 \% confidence value of chi-square test. 
Figure \ref{fig:gps_carpark} illustrates the rejection of GPS outliers when vehicle drives into an underground carpark, at which point the filter output relies only on dead reckoning. The GPS updates to the filter resume when exiting the carpark.
Figure \ref{fig:map} shows the resulting landmark map registration. Comparing to Figure \ref{fig:quadtoqueen_1} from method \ref{rigid}, it shows the usage of loose GPS edges can correct for drifts in addition to finding the landmark geographical positions.
Multiple loop closures further restricted the level of uncertainty in this dataset.

The global accuracy from this method can be variable and will be function of the quality of filtered GPS positions. For many applications such as the GPS-based initialization and relocalization using a feature map, this level of accuracy is sufficient. Nevertheless there are many urban vehicle application where higher level of accuracy are required that will mot be able to be assured when working with GPS based sensors in the typical city type environments.

\subsection{Fine tuning with anchor points in aerial imagery} \label{anchor}

\begin{figure}
    \centering
    \includegraphics[width=\columnwidth]{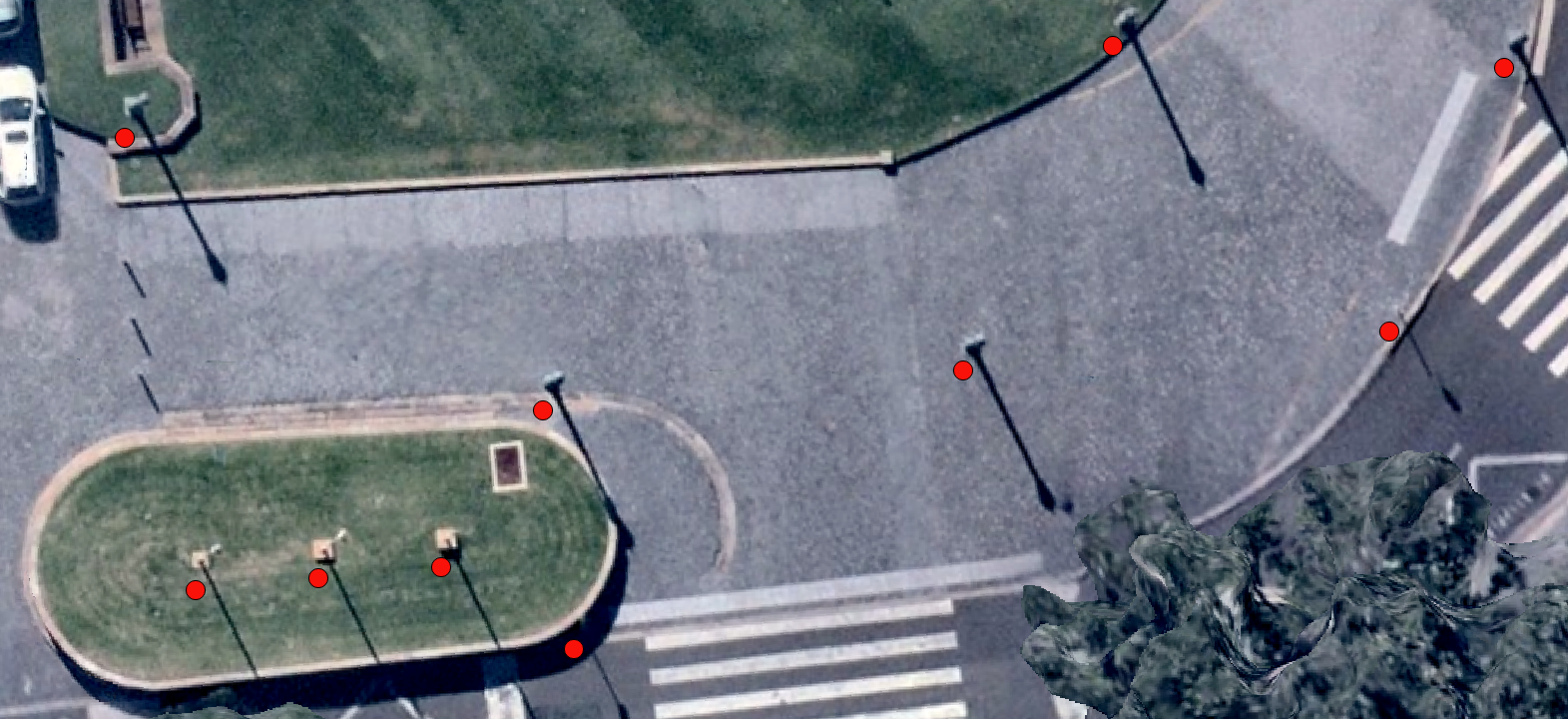}
    \caption{Close-up look of resulting landmark map from method \ref{loose}. Poles are clearly identifiable and can be accurately labeled. There is a 1.5-2m bias between lidar map and aerial image.}
    \label{fig:quad_closeup}
\end{figure}
\begin{figure}
    \centering
    \includegraphics[width=\columnwidth]{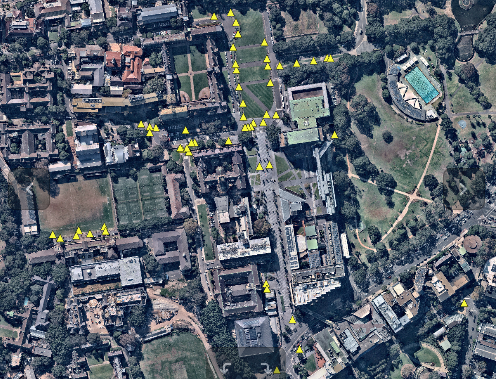}
    \caption{Positions of 50 labeled pole positions mainly concentrated in main quadrangle area. Only clearly visible poles are chosen and labeled as aerial map ground truth pole positions.}
    \label{fig:labeled_groud_truth}
\end{figure}
\begin{figure}
    \centering
    \includegraphics[width=\columnwidth]{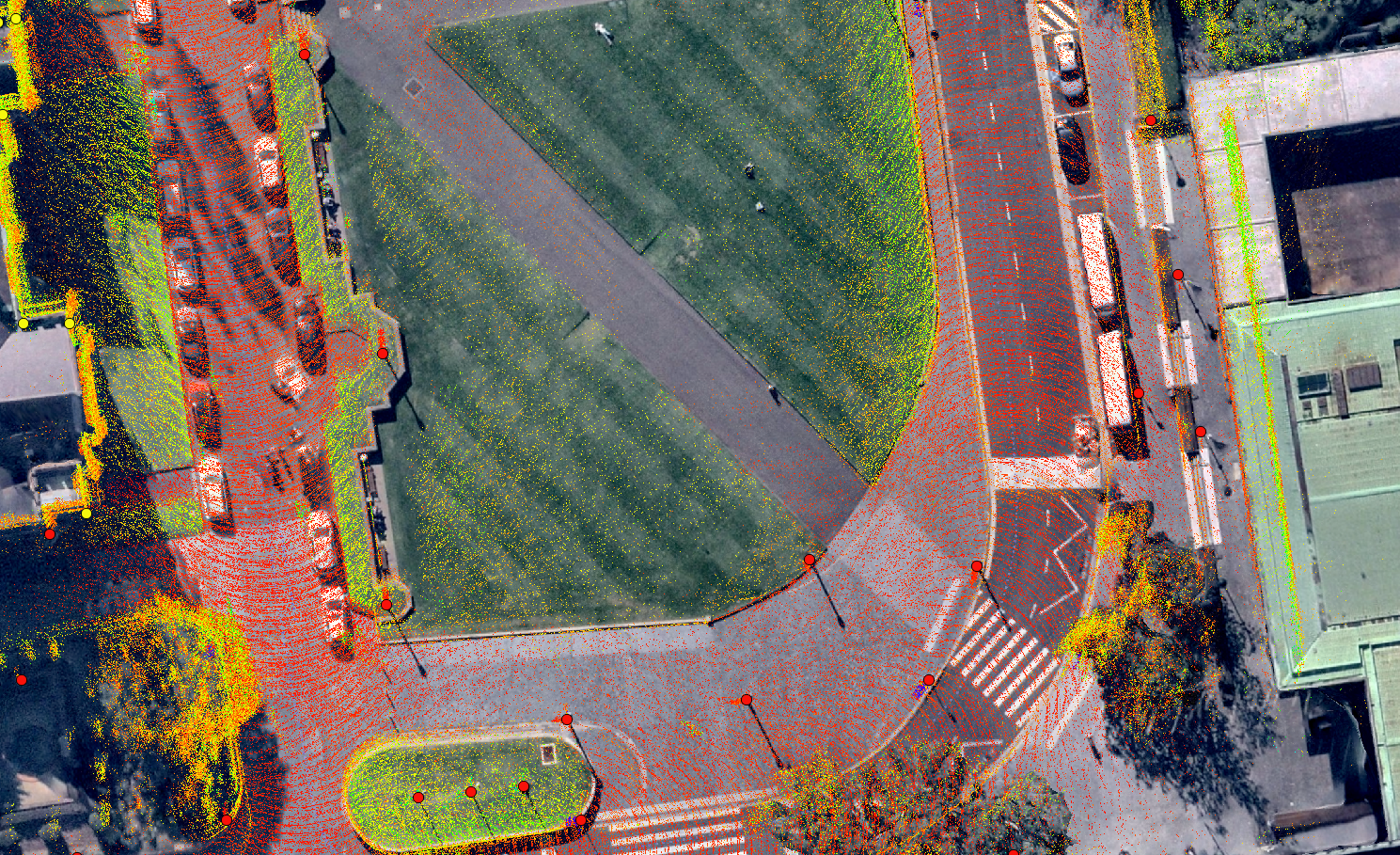}
    \caption{Close up look of accurately registered lidar orthoimage. Inaccuracies are mainly in longitudinal direction and when vehicle turning corners. Colors represent reflectivity.}
    \label{fig:quad_overlay}
\end{figure}

Aerial images are used in a variety of applications such as visualisation, mapping, surveying and extracting of features such as roads and buildings.
It is useful to link the feature map described in the previous sections to aerial imagery in order to find relationships between vehicles sensors and the top down aerial view.
The method described in the previous sections generates a georeferenced map with sufficient global accuracy for many applications. 
The registration methods described previously provided sufficient global accuracy for nearest neighbor data association of anchor points  but nevertheless, there are still biases and uncertainties with respect to the global ground truth.
This can be seen when the lidar features are projected into the aerial images, with an offset clearly visible as shown in Figure \ref{fig:quad_closeup}.
The bias in this figure could be attributed to either sensor noise in the GPS position information, or/and an offset in the registration of the aerial images.

We obtained from Nearmap\cite{Nearmap} an up-to-date orthogonal aerial images with absolute horizontal accuracy of 28-75cm, horizontal measurement accuracy of 11.5-15cm within one photo and 58-76cm between photos. This aerial orthoimagery has an 80\% higher resolution compared to standard satellite images such as from Google. Unlike standard satellite images, orthoimagery is strictly top-down view, meaning vertical building facades are lines instead of oblique faces, and building corners are points instead of lines. Figure \ref{fig:quad_closeup} shows the zoom view of pole landmarks obtained from method \ref{loose} and overlaid on Nearmap aerial image. Since the aerial picture resolution is high enough, the poles are clearly identifiable and have an 1.5-2m bias from lidar mapped poles. The correspondences can be confidently established through a nearest neighbor approach after the map is roughly registered by the strategy introduced in Section \ref{loose}. We labeled and obtained the position of 50 poles identified from the aerial imagery using a GIS capture tools as shown in Figure \ref{fig:labeled_groud_truth}. Only highly visible and clearly identifiable poles are labeled. We do not select poles where the visibility is compromised due to strong shadow, image blurs, blockage of trees and building rooftops. 

To minimise the error between the aerial images and the graph based feature map, we added additional graph edges corresponding to each of the 50 labelled poles using the image coordinates provided from the georeferences aerial image. 
The global graph optimisation was then carried out using all of the constraints from both local and global edges.
The edges joining the map origin to the aerial pole observations were added with a very low measurement uncertainty. This allows the graph to warp towards the hand labelled pole observations which in turn realigns the entire map.
To visualise the vehicle sensor data using this graph optimised trajectory, we next project all of the dense lidar pointclouds using the optimised vehicle poses. 
These pointclouds are then projected onto a horizontal plane and overlaid onto the aerial orthoimage as a GIS layer, as shown in Figure \ref{fig:full_overlay}. A close-up view in \ref{fig:quad_overlay} reveals that the lidar points are aligned very well in lateral direction (perpendicular to vehicle travel directions), but less accurate in longitudinal direction. This could be caused by motion distortion of lidar pointcloud and suggests future improvements using motion corrected pointclouds.

\section{QUANTITATIVE EVALUATION} \label{result}
\begin{figure}
    \centering
    \includegraphics[width=\columnwidth]{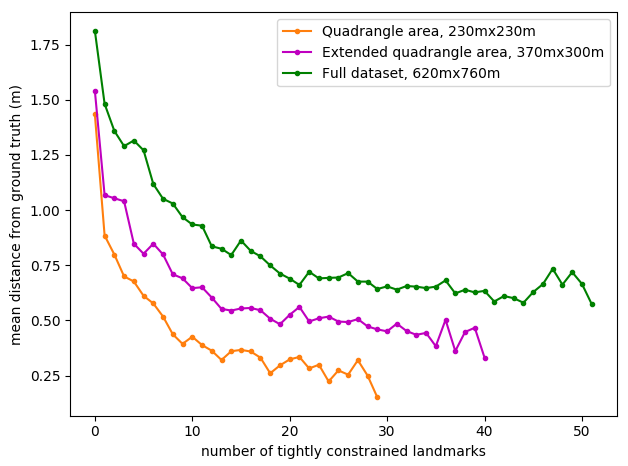}
    \caption{Absolute global accuracy evaluation. See section \ref{result} for detailed discussions. }
    \label{fig:dropout_lm}
\end{figure}

To evaluate the effect of increasing the number of tightly constrained aerial pole features on global map accuracy, we tied n=0,1,2,...49 lidar map poles to their corresponding aerial image global positions as shown in Figure \ref{fig:dropout_lm}. Next, we calculated the mean error/distance of the remaining (50-n) poles in the labeled pole set. 
Since we have only 50 highly identifiable poles in total, we are able to exhaustively evaluate all possible combinations of n constrained poles in the labeled pole set and compute the average of mean error of all combinations. When n=0 and no poles are constrained, the original average accuracy of method \ref{loose} output is 1.8m for 50 labeled poles in dataset. As more labeled poles are constrained, accuracy improved and plateaus around 57-75cm.

\begin{figure}
    \centering
    \includegraphics[width=\columnwidth]{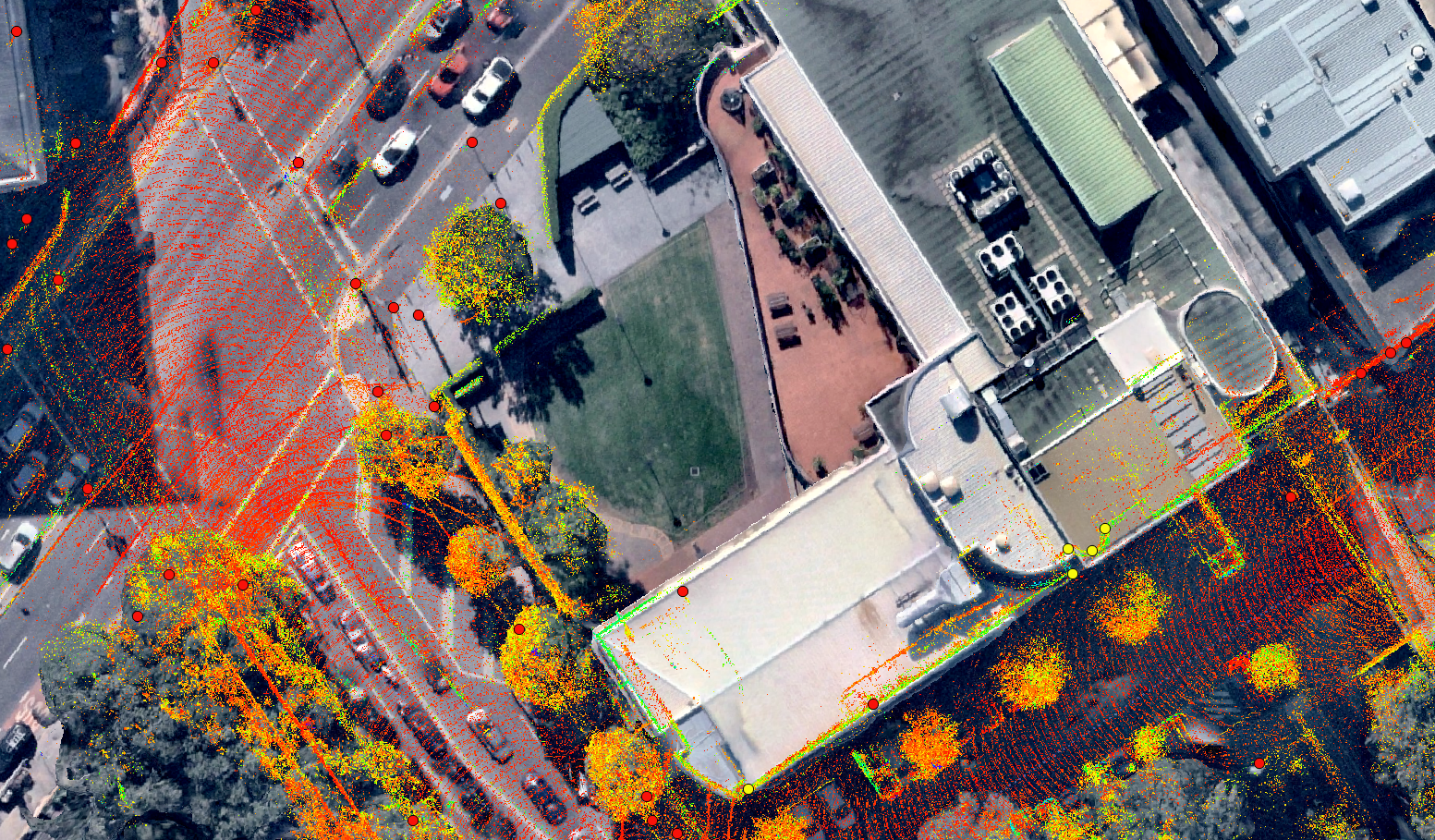}
    \caption{Lidar-aerial orthoimagery fusion. Red dots show positions of poles optimised from method \ref{anchor}. Visibility of most poles in this area is compromised therefore are not captured and constrained. Nonetheless we can speculate some of pole position from their strong shadows. Visual inspection confirms their registration accuracy is within 75cm. Line structures in lidar point clouds also accurately overlap with building walls, fences and bushes in aerial imagery. This is in an area far from quadrangle, where most of labeled ground truth poles are constrained. }
    \label{fig:overlay}
\end{figure}

Because the labeled poles are unevenly distributed and mostly concentrated around quadrangle area, we compare the results for both the quadrangle containing 30 labeled poles, and the extended quadrangle area which contains 40 labeled poles.
In quadrangle area, adding graph edges for any combinations of 29 poles result in a 15cm reduction in the global error when comparing to the full dataset area. Adding edges for 29 and 49 poles result in 64cm and 57cm mean errors respectively. 
This reduction in error is most likely a result of the combined aerial image being constructed from multiple individual images, and while the full dataset spans more than one image, the quadrangle area resides within a single image. 
Crossing image boundaries introduces additional error due to inconsistencies in the image stitching process. This is in accordance with the quoted aerial map horizontal measurement accuracy of 11.5-15cm within the same image and 58-76cm between images. It may prove to be a better strategy to optimize on smaller areas/sub-maps within each aerial image. Figure \ref{fig:overlay} shows the registration quality in a sector far from the quadrangle area when all 50 labeled poles are incorporated. It can be clearly seen that map landmarks and laser information have a very accurate match.






\section{CONCLUSIONS AND FUTURE DIRECTIONS} \label{conclusion}

In this paper, we presented a strategy to georegister a relative position graph-based feature map with globally registered aerial imagery. This was achieved by first loosely coupling the feature map to GPS positions in order to warp the feature map into a global reference frame with map quality constrained by the quality of the GPS information. The correspondences with the aerial imagery were created by manually labelling the features that could be observed in both the aerial images and the feature map. We demonstrate the improvement in map accuracy relative to the area surrounding the feature correspondences, and overall in the remainder of the map. The error in the map was reduced with the introduction of additional correspondences, approaching the quoted accuracy from the provider of the aerial images.  
Automatic detection of poles and other salient features in the aerial images will improve the scalability and accuracy of this method. 

Aerial oblique views and 3D reconstruction models enables richer feature extraction and make it possible to apply sensor fusion with ground survey sensors.
By mapping the local features into the coordinate frame of aerial imagery, lane and road markings extracted from the aerial images can be incorporated into the feature map, and correspondences between on board vehicle sensors can be generated.
These additional features can be used to improve the local and global accuracy of the maps for vehicle navigation. 

\bibliographystyle{ieeetr}
\bibliography{citation}

\addtolength{\textheight}{-12cm}   

\end{document}